\documentclass[journal]{IEEEtran}

\usepackage{cite}
\usepackage[hidelinks]{hyperref}
\usepackage[utf8]{inputenc}
\usepackage{graphicx}
\usepackage{amsmath}
\usepackage{amssymb}
\usepackage{amsthm}
\usepackage{booktabs}
\usepackage{algorithm}
\usepackage{algorithmic}
\usepackage{multicol}
\usepackage{multirow}
\usepackage{tabularx}
\usepackage{color}
\usepackage[switch]{lineno}

\begin{document}

%
\title{FDA-GAN: Flow-based Dual Attention GAN \\ for Human Pose Transfer}
%
%
%

\author{Liyuan~Ma,
        Kejie~Huang,~\IEEEmembership{Senior~Member,~IEEE},
        Dongxu~Wei,
        Zhaoyan~Ming,
        and~Haibin~Shen
\thanks{This work was supported by the National Natural Science Foundation of China (U19B2043). \textit{(Corresponding author: Kejie Huang.)}}
\thanks{L. Ma, K. Huang, D. Wei, and H. Shen are with the Department of Information Science \& Electronic Engineering, Zhejiang University, Hangzhou 310027, China(Email:mlyarthur@zju.edu.cn; huangkejie@zju.edu.cn; tracywei@zju.edu.cn; shen\_hb@zju.edu.cn).}
\thanks{Y. Ming is with Institute of Computing Innovation, Zhejiang University, Hangzhou 310027, China(Email:mingzhaoyan@gmail.com).}}


%
%

\maketitle

\begin{abstract}
Human pose transfer aims at transferring the appearance of the source person to the target pose. Existing methods utilizing flow-based warping for non-rigid human image generation have achieved great success. However, they fail to preserve the appearance details in synthesized images since the spatial correlation between the source and target is not fully exploited. To this end, we propose the Flow-based Dual Attention GAN (FDA-GAN) to apply occlusion- and deformation-aware feature fusion for higher generation quality. Specifically, deformable local attention and flow similarity attention, constituting the dual attention mechanism, can derive the output features responsible for deformable- and occlusion-aware fusion, respectively. Besides, to maintain the pose and global position consistency in transferring, we design a pose normalization network for learning adaptive normalization from the target pose to the source person. Both qualitative and quantitative results show that our method outperforms state-of-the-art models in public iPER and DeepFashion datasets.
\end{abstract}


%
\IEEEpeerreviewmaketitle

\section{Introduction}
%
%
%
%
\IEEEPARstart{H}{uman} pose transfer refers to the task of synthesizing human image with source human texture and target pose. We can utilize it to produce various human motions that are absent in the real world. This task has a wide range of multimedia and computer vision applications such as video meetings, virtual human generation, data augmentation for person re-identification, etc. 

\begin{figure}[!htb]
    \centering
    \includegraphics[width=\linewidth]{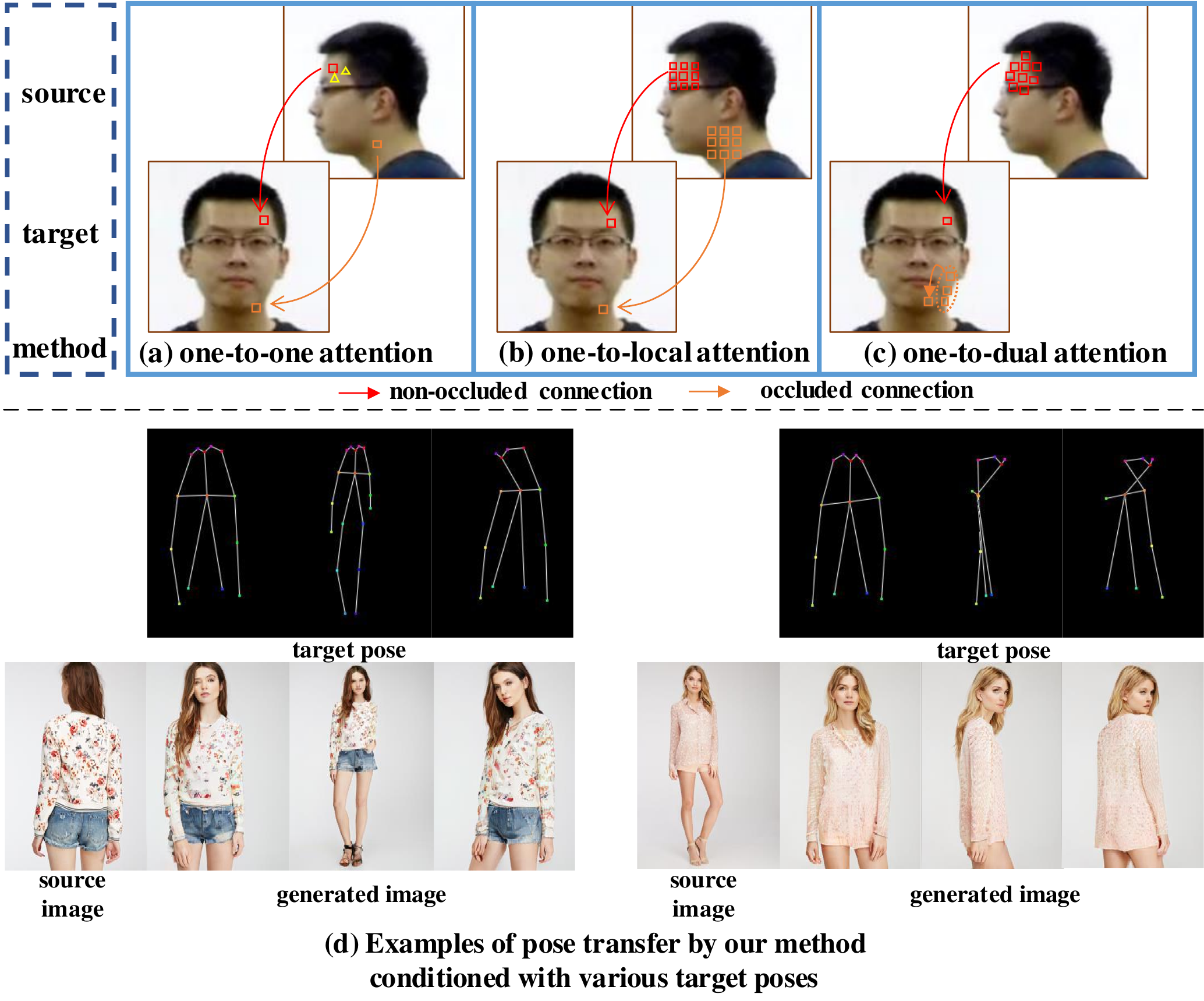}
    \caption{The illustration of our dual attention and the conventional sampling methods. The red and orange lines connect the attention locations with the invisible non-occluded target location and visible occluded target location. (a) The one-to-one attention builds the exclusive relationship between source and target, and the yellow triangles indicate the misleading values compared with the red square. (b) For one-to-local attention, each target value connects to a regular local source patch. (c) In our one-to-dual attention module, the non-occluded part of the target uses deformable blocks to sample the source's value externally. The occluded regions can internally relate themselves with target areas visible in the source. (d) We visualize the pose transfer results with our one-to-dual attention mechanism.}
    \label{fig:mapping}
\end{figure}

The generation of target-posed source image can be divided into two subtasks: reassembling source image parts coexistent in the target image and predicting the nonexistent parts to fill in the target image. By deforming the source image features to align them with the target pose, deformation-based methods like \cite{dong2018soft,siarohin2018deformable} have demonstrated great success. However, these methods are built on the rigid human body assumption and fail to model complex non-rigid body deformations for motion transfer tasks. The recent emergence of one-to-one \cite{li2019dense,liu2019liquid} and one-to-local \cite{ren2020deep} flow-based methods managed to address the issue mentioned above by predicting warping flows, which can establish point-wise correlations between source and target. However, the following weaknesses remain in these flow-based methods. As is shown in \autoref{fig:mapping} (a), the one-to-one mapping between source and target is sensitive to incorrect flow guidance caused by indistinguishable source values around the expected source position. The one-to-local attention further improves it by sampling a local source patch for each target position as visualized in \autoref{fig:mapping} (b). The local attention mechanism is still restricted by the limited receptive field, as the attention value is confined to a small local region. Notably, both methods cannot supply reliable attention values for the occluded parts without a source counterpart. Besides, most of the existing motion transfer methods cannot transfer target pose while retaining the source body size or global position. 

\begin{figure*}[!htb]
    \centering
    \includegraphics[width=\linewidth]{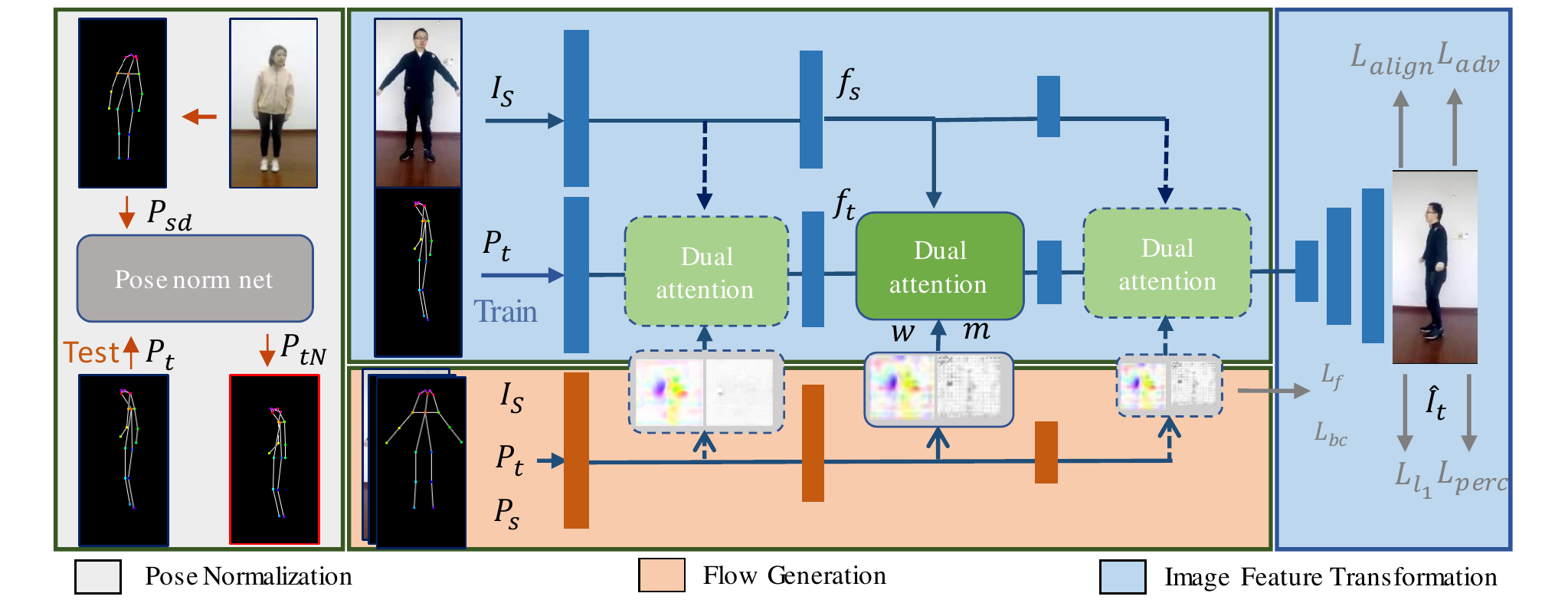}
    \caption{The overall framework of our FDA-GAN, which consists of three parts. The gray, orange, and blue blocks represent the Pose Normalization, Flow Generation, and Image Feature Transformation modules, respectively. Given the source image $I_s$ and pose pairs $(P_s, P_t)$, the Flow Generation module predicts the warping flow $w$ and occlusion map $m$. Then the dual-attention block propagates the source feature $f_s$ into the target feature $f_t$ in multi-feature level. Finally, it generates the target-posed image $\hat{I}_{t}$ with encoded target feature $f_t$. The Pose Normalization network is trained separately to provide consistent target pose input for Flow Generation and Image Feature Transformation in test period. It takes inconsistent pose pairs $(P_{sd}, P_t)$ as input, and normalizes the original target pose $P_t$ into the source-type pose $P_{tN}$.}
\label{fig:framework}
\end{figure*}
\indent This paper proposes Flow-based Dual Attention GAN (FDA-GAN) to attentionally integrate the source values with the target and infer the shape-invariant warping flow with adaptive pose normalization. Our one-to-dual attention module is composed of deformable local attention and flow similarity attention corporately. To minimize the impact caused by the incorrect flow estimation result, the deformable local attention samples a learnable irregular local patch around the corresponding source position rather than the fixed regular local patch. Therefore we can allow a larger receptive field for each target position and gain a more precise spatial correlation between source and target. Besides, we propose a bidirectional consistency loss to avoid sampling source's ambiguous attention values for occluded target positions. Thus we infer the occluded target values with the adjacent non-occluded positions that have similar warping flow values as shown in \autoref{fig:mapping} (c). In summary, the invisible target positions can be predicted from the visible ones with motion similarity and spatial correlation. \\
\indent When training pose transfer model on the video-based datasets like iPER\cite{liu2019liquid}, which collects consecutive posed human images for each person, the source and the target share the same pose structure in the training period. However, it becomes difficult for this model to handle pose pairs with high variance in testing, leading to undesired generated results. For consistent pose estimation, we utilize the smpl\cite{loper2015smpl} model to collect same-posed but different-shaped skeleton training samples and design the pose normalization network for learning the mapping. In the test, we transfer the target pose to keep it aligned with the source pose, which has no dependency on the estimated 3d vertices like \cite{liu2019liquid}.\\
\indent To prove our model's superiority, we compare it with other state-of-the-art methods performing image-based human pose transfer. The experiment results qualitatively and quantitatively show that our model outperforms the conventional schemes, especially in image details and human appearance consistency.  Moreover, we conduct ablation studies and prove its effectiveness with detailed analysis.

In general, we summarize our contributions as follows:
\begin{itemize}
\item We introduce a feedforward network to estimate the bidirectional consistent 2d warping flow, which uses forward-backward mapping to check the misestimated flow values brought by self-occlusion or fuzzy texture.
\item We propose a flow-guided dual attention block to deform and reassemble the critical image features into the generated result, which works in the generation of the occluded regions by flow similarity measurement.
\item We propose the pose normalization network to align better the source and the target pose, which can help to generate the target posed human of similar build and body structure with the source human.
\end{itemize} \leavevmode

The rest of this paper is structured as follows: Sec.\ref{section:rel} introduces the related works. Sec.\ref{section:method} describes the details of our proposed FDA-GAN. The analysis of our results is presented in Sec.\ref{section:exp}. In Sec.\ref{section:discussion} we discuss our limitation and the future directions for improvement. Finally, the paper is concluded in Sec.\ref{section:conclusion}.

\section{Related Work}
\label{section:rel}
\subsection{Person Image Synthesis}
The development of image-based pose transfer is largely based on the flourishment of image synthesis techniques. Unlike the conventional synthesis methods, which heavily rely on the dedicated hand-crafted features, the emergence of Generative Adversarial Networks (GANs)~\cite{goodfellow2014generative, isola2017image, johnson2016perceptual, ledig2017photo, mirza2014conditional, odena2017conditional, radford2015unsupervised, yu2019free, 9141513, liu2017face, shu2020host} brings new insight to the sharp image generation by two-players adversarial learning. Much of the current literature pays attention to generating images under some conditional constraints, e.g., desired pose, viewpoint, sketch, etc.  Conditional GANs (cGANs)\cite{mirza2014conditional} have achieved impressive performance in controllable image generation. \cite{chen2019quality} demonstrates their remarkable scalability to handle problems such as sketch-to-image, label-to-image. However, the pose-conditioned image transfer cannot easily be tackled by pixel-wise aligned image transfer methods due to the unaligned nature between pose pairs. \cite{lassner2017generative} introduces a model to combine the variation auto-encoder and GAN to synthesize the vivid person images, which can generate person images with different appearances. \cite{balakrishnan2018synthesizing} splits the person image synthesis task into the generation of foreground and the background.  \cite{yang2020towards} proposes a model for synthesizing the fashion images conditioned with pose map and textual description. \cite{liu2019swapgan} presents a unified multi-stage deep generative model to tackle the multi-conditional person image generation. \cite{men2020controllable} designs a novel generator architecture with attribute decomposition and recombination to handle attributed-guided person synthesis under various appearances and poses. \cite{lu2021generate} generates the person images for data augmentation, which uses body-part maps to attentionally entangle the appearance and structure features. \cite{kim2019style} proposes an approach to characterize the person clothing segments with disentangled geometry and style modeling, generating a realistic-looking new-clothed person with fine-level style control.

\begin{figure}[!tb]
\centering
\includegraphics[width=\linewidth]{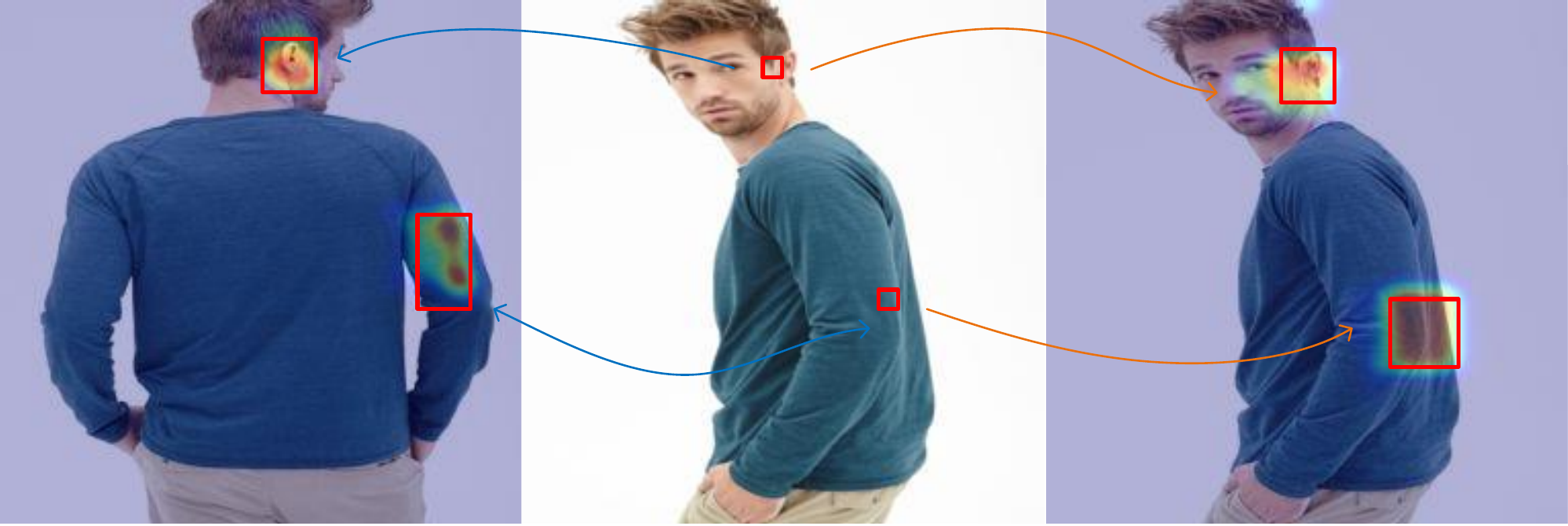}
    \caption{The visualization of the dual attention heatmaps. Images from left to right are: the source input, the generated result, and the corresponding encoded target. The left blue line and the right red line indicate the sampling locations with our deformable local attention module and flow similarity attention, respectively. The attention coefficients are visualized with heatmaps.}
\label{fig:dualattn}
\end{figure}

\subsection{Human Pose Transfer}
Human pose transfer is a more challenging task. \cite{ma2017pose} is the first work to deal with human pose transfer. It uses a two-stage U-Net to transfer the target person’s pose to the reference image. After that, the generated images are refined with sharper texture in an adversarial way. The subsequent works \cite{tulyakov2018mocogan,ma2018disentangled} further split the task into pose generation and appearance generation to avoid disturbance from the other. However, the above Encoder-decoder methods all fuse the motion and content information in an add or concatenate manner, taking no deformation between paired poses into consideration. To learn the deformation in different poses, \cite{song2019unsupervised, dong2018soft} utilize the result of human parsing \cite{gong2018instance} as ground truth to learn the spatial transformation in feature and image level, respectively. \cite{siarohin2018deformable} and \cite{zhou2019dance} treat the transformation as a set of body parts affine transformation. \cite{siarohin2018deformable} proposes deformable affine skip connection in U-Net to define the local feature transformation for each body part based on the rigid body assumption. Unlike the above one, \cite{zhou2019dance} performs the human pose transfer in the image level and recomposes body part images to generate the new posed image.

Nonetheless, these methods commonly conflict with human's non-rigid nature and thus limit their performance on the task. With the help of a 2d flow map or a 3d corresponding map, \cite{wei2019gac,wei2020c2f,liu2019liquid,li2019dense,zheng2020pose} try to extract the reliable appearance information with multi-scale flow and reach higher warping performance. \cite{knoche2020reposing,ren2020deep,tang2021structure} realize that warping source at the pixel level prevents the model from generating new content. They perform feature deformation to source features, which can better propagate the source information to the composed target in the feature space. Besides, most of the existing motion transfer methods cannot transfer target poses while retaining the source body sizes or global positions. Even if some recent methods \cite{liu2019liquid,li2019dense} can alleviate this by describing motions with 3d human models rather than 2d poses, they deeply rely on precise reconstructed 3d models and thus frequently fail.

\subsection{Attention Mechanism in human pose transfer}
Attention mechanism is widely used to assist the feature matching~\cite{shu2021spatiotemporal}. In motion transfer, an attention mechanism can help the model to find more reasonable sampling positions. \cite{zhang2019attention,zhu2019progressive,horiuchi2019spectral,li2020pona} propose a pose-guided attention network to avoid misalignment. However, their self-attention module has a high computational burden and considers excessive unrelated information. Flow-based methods regard target image as the deformation of source image, \cite{ren2020deep} propose a local attention framework to calculate each output position with a local source patch whose center is provided by feature flow. The local attention can avoid the poor gradient brought by bilinear sampling, and we extend this idea to use a deformable local patch. Our learnable deformable kernel can capture more relevant information around the patch center.

\begin{figure}[!tb]
\centering
\includegraphics[width=\linewidth]{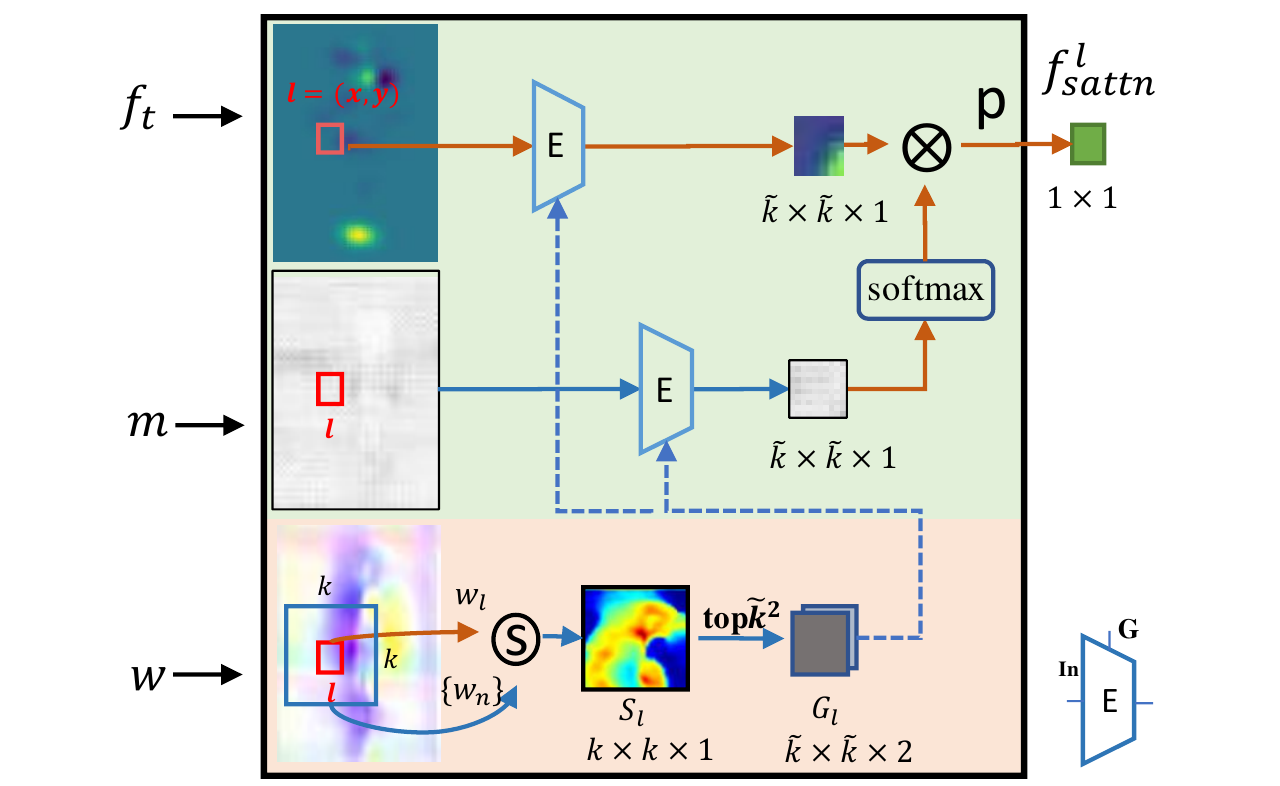}
    \caption{The illustration of our flow similarity attention module. For readability, we take the flow similarity computation process of location $l(x,y)$ as example and then apply the calculation for every location in the target feature map. The extractor \textbf{E} represents the bilinear grid sampler which takes \textbf{In} as input and \textbf{G} as grid index map, and the \textbf{S} means the flow similarity calculation block.}
\label{fig:fa}
\end{figure}

\section{Method}
\label{section:method}
FDA-GAN consists of three parts: \textbf{Flow Generation}, \textbf{Image Feature Transformation}, and \textbf{Pose Normalization}. As is shown in \autoref{fig:framework}, the Flow Generation module generates an bidirectional consistent feature flow $w$ to obtain valid sampling locations from the source appearance feature. Based on such sampling constraints, the Image Feature Transformation module utilizes our dual-attention (i.e., \textit{flow similarity attention} and \textit{deformable local attention}) blocks to sample useful appearance features. Then it takes the encoded target features as inputs, and synthesizes the final output image $\hat{I}_{t}$ with a decoder. Moreover, considering the substantial skeleton structure difference between source and target when they belong to different individuals, we utilize Pose Normalization to minimize the structure disparity between source and target skeletons.

\subsection{Bidirectional Consistent Flow Generation}
The flow generation module aims to generate warping flows that align encoded source features with the target features. The warping flows are used to place the source feature values in the target feature map. The pretrained VGG \cite{simonyan2014very} can supply image's spatial distribution information in multi-feature level. Thus with the VGG features of source image warped by flow $\mathbf{w}^f$ and ground truth target images, we can predict the $\mathbf{w}^f$ by minimizing the flow loss at all N positions in the $\Omega$ coordinates set of feature maps to learn the mapping relationship:
\begin{equation}
\mathcal{L}_{f}=\frac{1}{N} \sum_{l \in \Omega} \exp \left(-\phi\left(\mathbf{vgg}_{s, \mathbf{w}^f}^{l}, \mathbf{vgg}_{t}^{l}\right)\right)
\end{equation}
where $\mathbf{vgg}_{s, \mathbf{w}^f}^l$ and $\mathbf{vgg}_{t}^l$ represent the VGG feature values of warped source and ground truth target located in $l$, and $\phi(\cdot)$ means the cosine similarity function.

However, the estimated flow merely uses forward mapping to supervise the flow learning, which is sensitive to perturbation from indistinguishable matching pairs as shown in \autoref{fig:mapping} (a). We then take a forward-backward mapping check with bidirectional consistency loss to better differentiate the analogous feature positions and mark the occluded ones in the occlusion mask. The motivation comes from that the sum of the forward flow and the backward flow at the corresponding non-occluded positions should be zero to achieve the correct mapping relationship. Correct bilateral matching between the source and the target distinguishes occluded and non-occluded positions with occlusion mask $m$:
\begin{multline}
    \mathcal{L}_{{bc}}= 
    \sum_{l \in \Omega}m_{l}^{f} \cdot \rho\left(\mathbf{w}^{f}(l)
    +\mathbf{w}^{b}\left(l+\mathbf{w}^{f}(l)\right)\right) \\
    \quad+m_{l}^{b} \cdot \rho\left(\mathbf{w}^{b}(l)+\mathbf{w}^{f}\left(l+\mathbf{w}^{b}(l)\right)\right)
\end{multline}

The forward flow $\mathbf{w}^f$ and backward flow $\mathbf{w}^b$ are estimated by computing them from both flow directions (i.e.,source to target and target to source), along with the occlusion masks $m^f$ and $m^b$. $\rho(\cdot)$ are the robust Charbonnier function $\rho(x) = {\left( x^2+\epsilon^2\right)}^{K}$.

\begin{figure}[!tb]
\centering
\includegraphics[width=\linewidth]{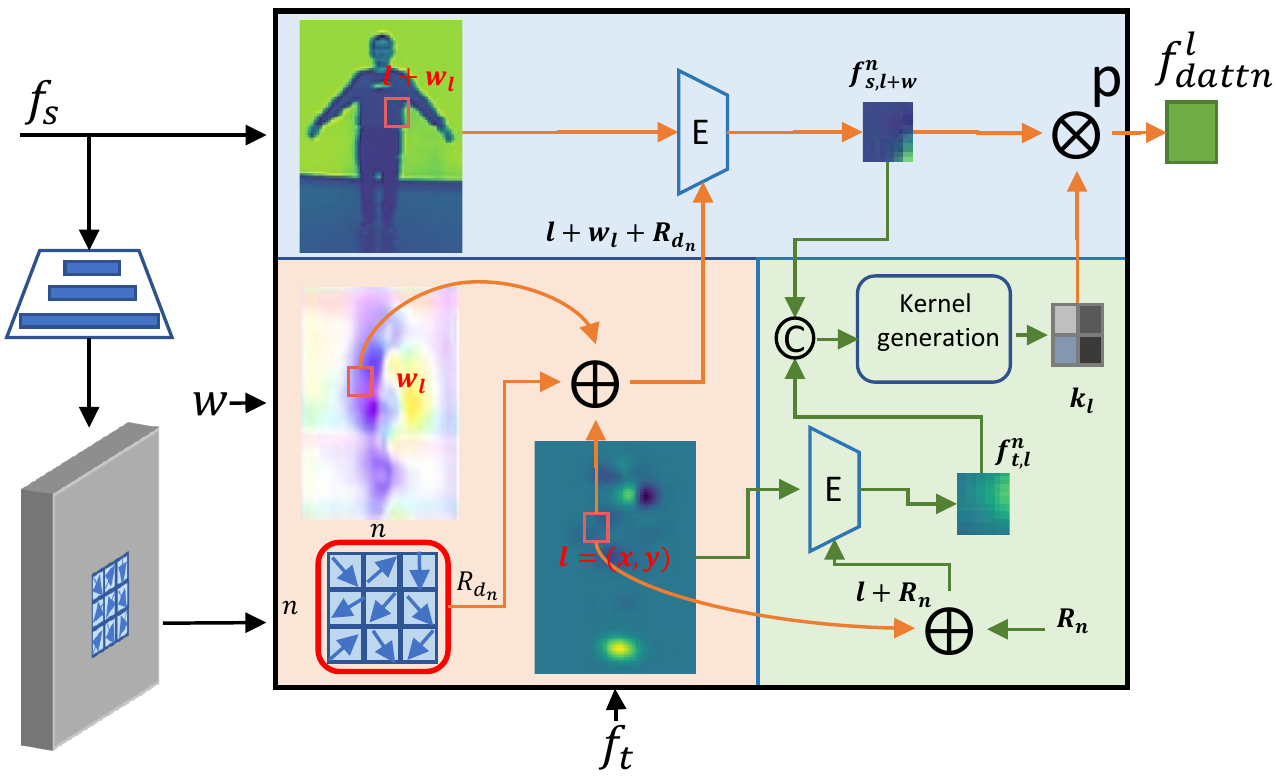}
    \caption{The illustration of our deformable local attention module.}
\label{fig:da}
\end{figure}

\subsection{Dual Attention Image Feature Transformation}
\label{section:dualattn}
Our image feature transformation module is based on the dual attention mechanism. As shown in \autoref{fig:dualattn}, the dual attention module consists of the flow similarity attention block and the deformable local attention block, which are used to sample relevant positions from the encoded target and the source, respectively.

\subsubsection{Flow Similarity-based Attention Block}
The flow generation module aims to warp the source features to align with the target features. However, due to inconsistency with the source, some invisible target parts may be assigned with inappropriate source features during warping, degenerating the results. Our flow similarity-based attention block is inspired by the fact that similar appearances are more likely to have similar motions (denoted by flow similarity in our method). It samples target features visible in the source instead of ambiguous source features for these target parts, where the sampled target features have high flow similarity.

We group regions with the similar motion by measuring the direction and displacement coherence of flow vectors with the cosine similarity kernel $K_{c}$ and the gaussian kernel $K_{g}$, respectively. As is shown in \autoref{fig:fa}, given the flow value $w_l$ located in $l(x,y)$, the similarity evaluator \textbf{S} calculates the similarity map $S_l^n$ by comparing $w_l$ with its neighbors $\{w_n|n \in \mathcal{N}_{k}\left( l\right)\}$.
\begin{equation} \label{eq1}
\begin{split}
    S^n_l & =K_{c}\left(w_{l}, w_{n}\right) + K_{g}\left(w_{l}, w_{n}\right)\\
     & = \alpha \frac{w_{l}^{T} w_{n}}{\left\|w_{l}\right\|_{2}\left\|w_{n}\right\|_{2}}+ 
        \beta\exp \left(-\frac{\left\|w_{l}-w_{n}\right\|_{2}^{2}}{2 \sigma^{2}}\right), 
        n \in \mathcal{N}_{k}\left( l\right)    
\end{split}
\end{equation}
where $w_n$ comes from $k \times k$ neighbors $\mathcal{N}_{k}\left( l\right)$ around the $l(x,y)$ rather than the whole flow map to reduce the calculation cost and filter out the remote irrelevant values. The $\alpha$ and $\beta$ represent the respective weights for both kernels and the gaussian kernel bandwidth is set with $\sigma = 0.06$.

Afterwards, we calculate ${\Tilde{k}}^2$ largest elements's indexes from $\{S^n_l|n \in \mathcal{N}_{k}\left( l\right)\}$, and reshape them to $\Tilde{k}\times \Tilde{k}$ sampling index map $G_l:\{G^i_l|i \in \mathcal{N}_{\Tilde{k}}\left( l\right)\}$. The extractor \textbf{E} samples flow similarity attention values and weight values from target feature map $f_t$ and occlusion mask $m$. The occlusion map shows the probability of being occluded or not. Thus our attention module can utilize the feature values tending to be non-occluded to calculate the flow similarity attention value.\\
\indent After that, we normalize weight maps by softmax function, ensuring the reliability of gradient propagation. Eventually, the flow similarity attention result at location $l=(x,y)$ is computed as: 
\begin{equation}
    f_{sattn}^l = P\left({E(f_t,G_l)\otimes softmax(E(m,G_l))}\right)
\end{equation}
where $\otimes$ means the element-wise multiplication, and ${P}$ represents the average pooling. We can get the flow similarity attention values $f_{sattn}$ by traversing through every locations.

\subsubsection{Deformable Local Attention Module}
The original local attention sampling strategy uses a regular grid $\mathcal{R}^n$ over the source feature map $f^{l+w}_s$ to extract source feature patches around $l+w$. The grid $\mathcal{R}^n$ defines an $n \times n$ sampling kernel:
\begin{equation}
\mathcal{R}^n=\{(-\frac{n-1}{2},-\frac{n-1}{2}), \ldots,(\frac{n-1}{2},\frac{n-1}{2})\}
\end{equation}

However, the predefined square sampling region restricts its performance since the adjacent feature values may not supply correct sampling values. As shown in the \autoref{fig:da}, we augment the regular grid $\mathcal{R}^n$ by predicting the $n \times n$ irregular sampling kernel for each source sampling location $l+w$ in deformable local attention block. For literal simplicity, we denote it as $\mathcal{R}_{d_n}$. Then we get the deformable local attention sampling index map by summing the offset map $\mathcal{R}_{d_n}$ with flow value $w_l$ and original location $l$. After that, the extractor \textbf{E} calculates the $n \times n$ sampled block $f_{s,l+w}^n$ in the source map corresponding to target location $l$. The sampling result $f_{t,l}^n$ in target feature map is acquired by regular sampling centered in target position $l$. The kernel generation block takes the concatenation of $f_{t,l}^n$ and $f_{s,l+w}^n$ as input, followed by convolution blocks and softmax function to output the $n \times n$ weighting values $\mathbf{k}^n_l$ for corresponding sampled source feature block. The deformable local attention value for target location $l$ is defined as:

\begin{equation}
 \mathbf{f}_{dattn}^l=\frac{1}{{n}^2}\sum_{i \in \mathcal{R}^n,j \in \mathcal{R}^{d_n}} {k}^n_l\left(i\right) \cdot \mathbf{f}^n_{s,l+w}\left(j\right)
\end{equation}

The offset value should be constrained within the reasonable region. A $\ell_{2}$ norm sampling alignment loss is thus proposed to lower the offset map's variance around position $\left(l+w\right)$ of source feature maps.

\begin{equation}
    \mathcal{L}_{align}=\sum_{l \in \Omega}\sum_{j \in \mathcal{R}_{d_n}}\left\|f_{s,l+w} -f^j_{s,l+w} \right\|_{2}^{2}
\end{equation}
\indent In summary, using the estimated occlusion mask $m$, we selectively combine the warped source feature $f_{sattn}$, flow similarity attention result $f_{dattn}$, and the target feature $f_t$ to get the final generated feature map:
\begin{equation}
\mathbf{f}_{{g}}=(1-m) * (f_t+f_{sattn})+m * f_{dattn}
\end{equation}
where the deformable local attention value $f_{dattn}$ warps the non-occluded source feature maps and the occluded parts are predicted with $f_{sattn}$. 

Apart from the losses mentioned above, we further use several losses to train the network, namely adversarial loss, perceptual loss and reconstruction loss. Our adversarial loss adopts a vanilla GAN to approximate the distribution between generated result and ground truth, which is defined as:

\begin{equation}
\mathcal{L}_{a d v}=\mathbb{E}\left[\log \left(1-D\left(G\left({I}_{s}, {p}_{t}, {w}, {m}\right)\right)\right)\right]+{E}\left[\log D\left({{I}}_{t}\right)\right]
\end{equation}

To stabilize the training process and reduce the reconstruction errors, we also apply an L1 constraint between the generated image and the ground truth:

\begin{equation}
\mathcal{L}_{\ell_1}=\left\|\hat{{I}}_{t}-{I}_{t}\right\|_{1}
\end{equation}

The pretrained extracted activation maps are used to penalize in the feature level. The perceptual loss is calculated as:

\begin{equation}
\mathcal{L}_{\text {perc}}=\sum_{j=1}^{N}\left\|\phi_{j}\left(\hat{I}_{t}\right)-\phi_{j}\left(I_{t}\right)\right\|_{2}^{2}
\end{equation}

where $\phi_{j}$ means the $i\text{-th}$ layer of the pretrained VGG network. Our model is trained by a weighted sum of losses:

\begin{equation}
\begin{array}{r}
\mathcal{L}_{G}=\lambda_{1} \mathcal{L}_{f}+\lambda_{2} \mathcal{L}_{bc}+\lambda_{3} \mathcal{L}_{\ell_{1}}+\lambda_{4} \mathcal{L}_{\text{align}} \\
+\lambda_{5} \mathcal{L}_{a d v}+\lambda_{6} \mathcal{L}_{\text {perc}}
\end{array}
\end{equation}

\begin{figure}[!tb]
\centering
\includegraphics[width=\linewidth]{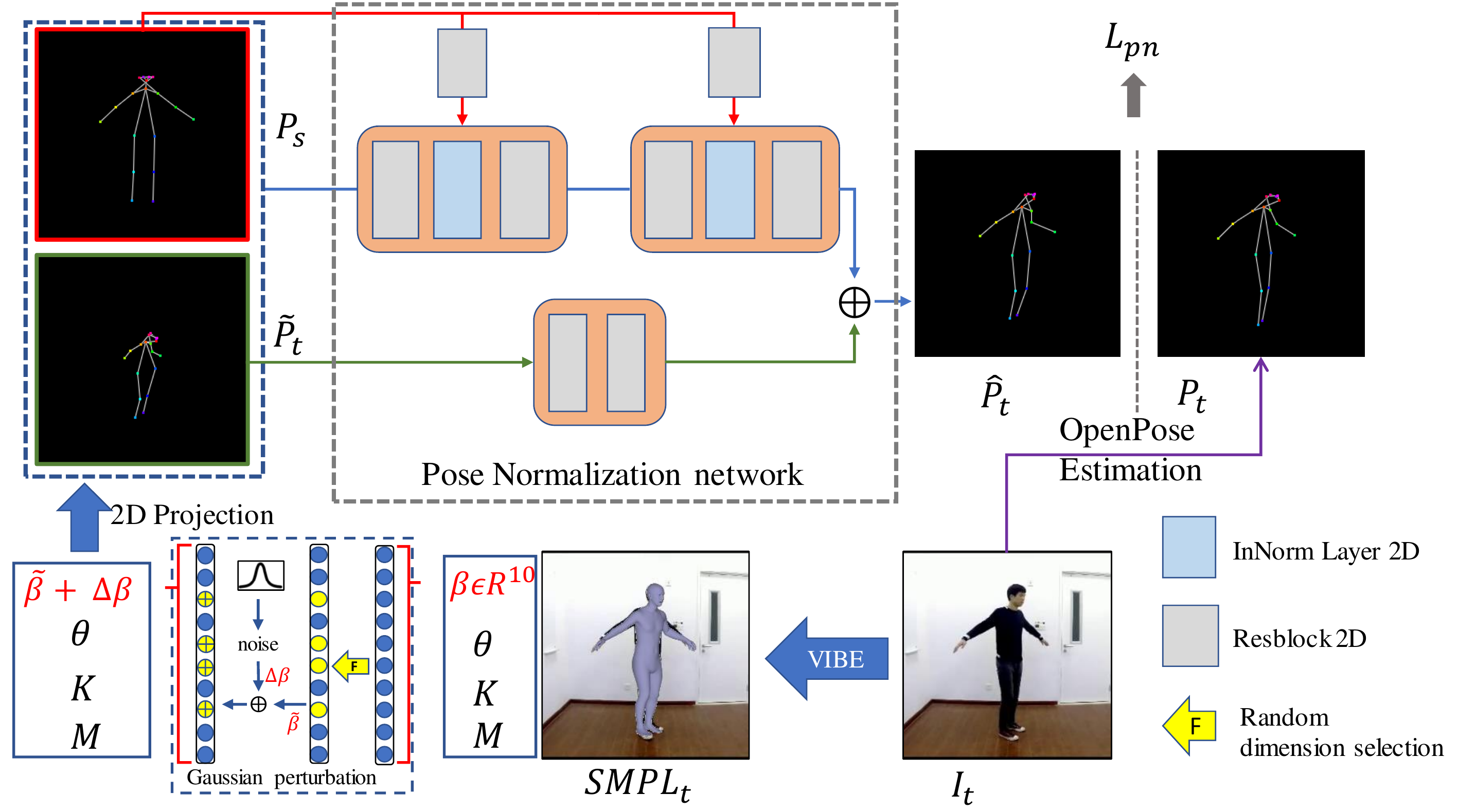}
\caption{The training pipeline of pose normalization network. The blue and gray block in the right bottom indicate the 2d instance normalization layer and residual convolution blocks which are the main components of our pose normalization network. The yellow circles represent shape param's dimensions applied with gaussian perturbation.}
\label{fig:posenorm}
\end{figure}

\subsection{Self-supervised Pose Normalization}
Motion transfer aims to preserve the source person's appearance, global location, and body shape. Since ground truth skeleton with the same pose but different shape and global location is scarce by nature, we address this problem by fitting SMPL\cite{loper2015smpl} model with 3d body model parameters estimator \cite{kocabas2020vibe} for target human. The SMPL model represents the human body with parameters including pose $\theta \in {R^ {N^{72}}}$, shape $\beta  \in {R^{10}}$, weak-perspective camera $K \in {R^3}$, and the render function $M(\theta, \beta) \in {R}^{6890 \times 3}$. The shape param $\beta \in R^{10}$ controls different aspects of body shape like body height, proportion, etc. Thus, the body shape can be altered by adding gaussian perturbation to specified dimensions as is shown in \autoref{fig:posenorm}. Then we project the new-shaped 3d body into 2d plane to get pose map $\Tilde{P}_t$. In the testing period, unlike the LiquidNET\cite{liu2019liquid} which needs an extra 3d human model to generate the target pose by deforming the source pose, our pose normalization network shows no dependency on costly 3d modeling.

\autoref{fig:posenorm} reveals that $\Tilde{P}_t$ has a different shape and global position but same pose with target pose map ${P}_t$. To normalize skeletons into source type, we use $\Tilde{P}_t$ as input and design the self-supervised pose normalization network to reconstruct ${P}_t$. Eighteen joint heatmaps define the input and output pose maps, so the widely used cross-entropy loss is applied to minimize the gap between model output $\hat{P}_t$ and ground truth ${P}_t$.

\begin{equation}
\mathcal{L}_{pn}=-\sum_{i=1}^{18} {P}^i_t \log \left(\hat{P}^i_t\right)
\end{equation}

The pose normalization network architecture is inspired by \cite{pavllo20193d}. We modify it with 2d convolution layers, instance norm layers, and residual structure. The instance norm layers are utilized to diminish the structure difference between $P_s$ and $\Tilde{P}_t$. Further, the residual structure can capture the structure residuals caused by varying shapes.

\begin{figure*}[!tb]
\centering
\includegraphics[width=\linewidth]{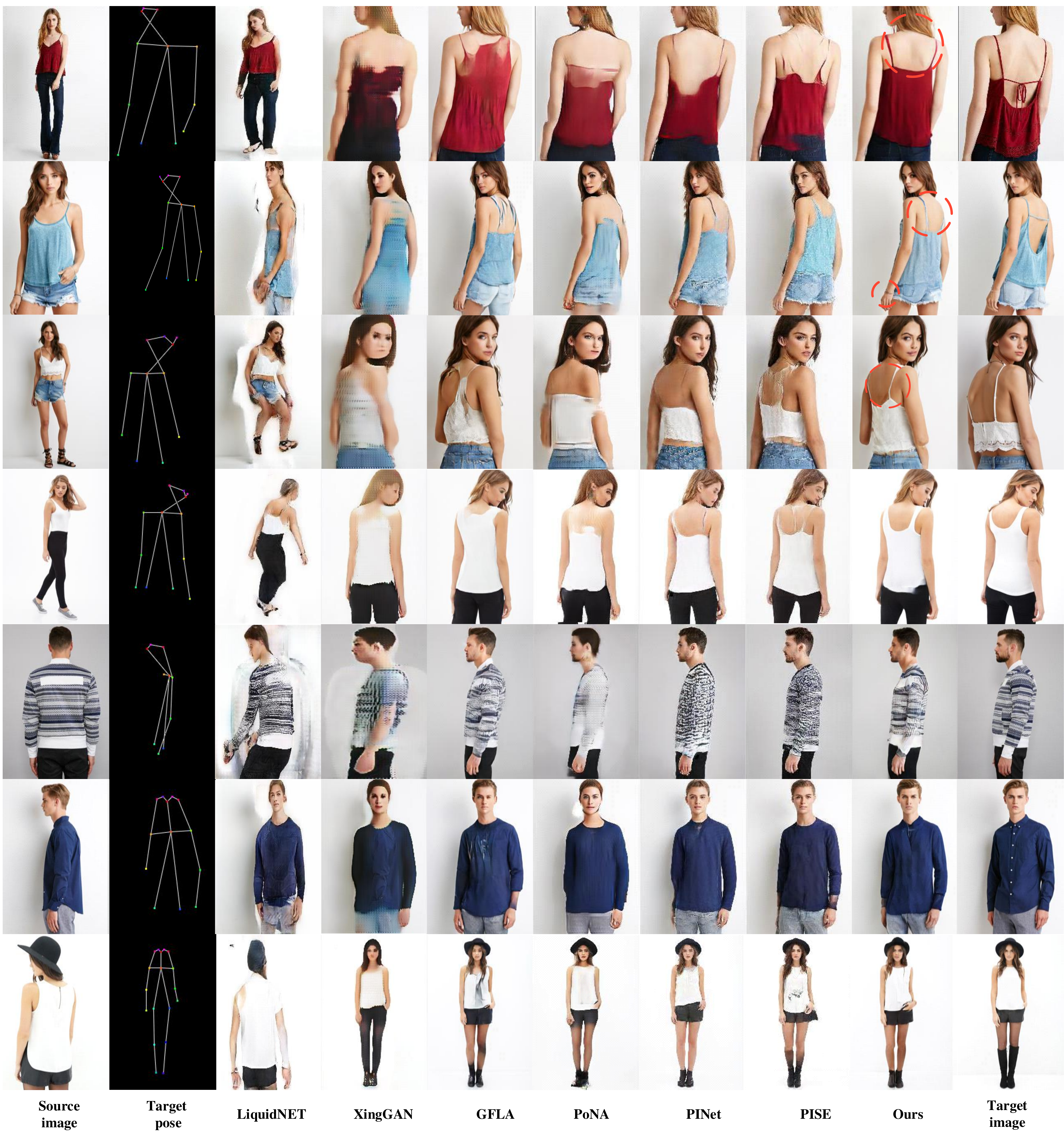}
\caption{Qualitative comparison in DeepFashion dataset between our model and the conventional methods (please zoom in for better view). From left to right are: input source image, input target pose, results of LiquidGAN\cite{liu2019liquid}(2019), results of XingGAN\cite{tang2020xinggan}(2020), results of GFLA \cite{ren2020deep}(2020), results of PoNA\cite{li2020pona}(2020), results of PINet\cite{zhang2020human}(2020), and results of PISE\cite{zhang2021pise}(2021).}
\label{fig:comparison_fashion}
\end{figure*}

\begin{figure*}[!tb]
\centering
\includegraphics[width=\linewidth]{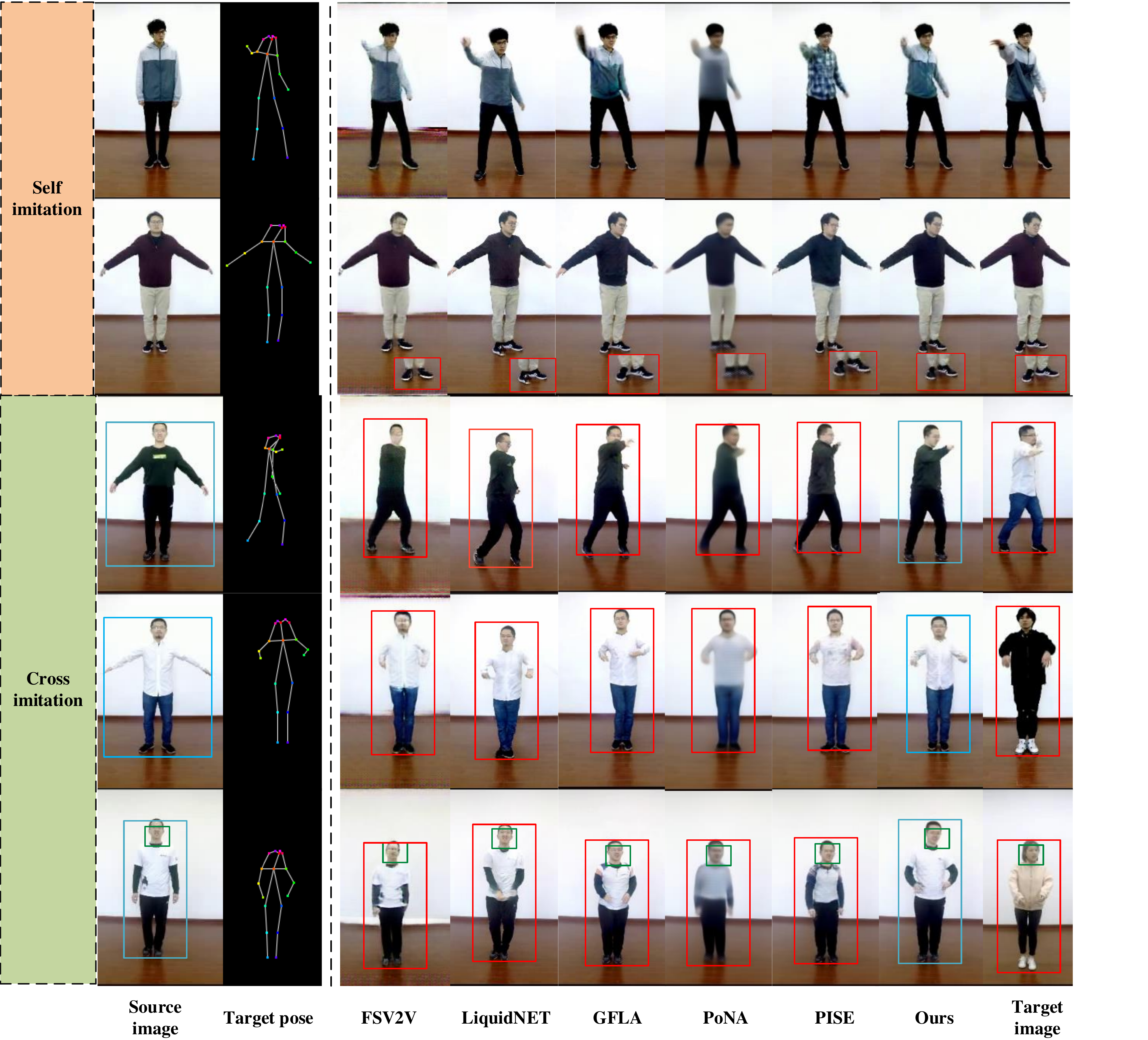}
\caption{Qualitative comparison between our model and the conventional methods including FSV2V\cite{wang2019fewshotvid2vid}(2019), LiquidGAN\cite{liu2019liquid}(2019), GFLA\cite{ren2020deep}(2020), PoNA\cite{li2020pona}(2020), and PISE\cite{zhang2021pise}(2021). We emphasize our model's better rendering of occluded region with enlarged shoes image. The red and blue boxes show the contrast of inability and ability to maintain the source human body shape and global position. The green box shows the comparison of different local pose (head) adjustment. {Best viewed enlarged on screen}.}
\label{fig:comparison_iper}
\end{figure*}


\section{Experiment}
\label{section:exp}
In this section, we first introduce the datasets, the training details, and evaluation metrics in Sec.\ref{section:di}. Then we compare our method with conventional methods in Sec.\ref{section:comparison}. Finally, we analyse the efficiency of the FDA-GAN Framework and the pose normalization network in Sec.\ref{section:effect_model} and Sec.\ref{section:effect_pn}.

\subsection{Implementation Details}
\label{section:di}
\subsubsection{Datasets}
In the task of human motion transfer task, we use video-based dataset iPER\cite{liu2019liquid} and image-based dataset DeepFashion In-shop Clothes Retrieval Benchmark \cite{liu2016deepfashion} to evaluate the performance of our model. The iPER dataset contains 206 high-resolution video sets from 30 persons. The human subjects in iPER videos are filmed with static viewpoints and show various motions. The DeepFashion dataset contains 72712 high-fidelity images with dynamic viewpoints, varying clothes, and background. Both are challenging in the scope of human pose transfer. We extract the human key points with OpenPose \cite{cao2019openpose} to get the pose information. Further, we split the iPER dataset into 164 training videos and 42 testing videos and collect 101966 training image pairs and 8570 testing image pairs for the DeepFashion dataset. To ensure the generalization ability of our method, the person identities in the training set do not overlap with the testing set.
\subsubsection{Network Architecture and Training Procedure}
We employ a triple-encoder with a single decoder architecture as the generation network in our experiment. All three encoders share the same network structure with three times downsampling. The initial search area size $k$ and ${\Tilde{k}}$ of $top{\Tilde{k}}^2$ operation are set to 10 and 4 for calculating the flow similarity attention, respectively. The whole training process is divided into two stages. First, we train the pose transfer model end-to-end with the estimated flow map and occlusion map. Then we separately train the pose normalization model as described in Sec.\ref{section:effect_pn} and apply it before the pose transfer model in testing. The overall training period costs ten epochs with Adam optimizer (learning rate:$2\times10^{-4}$) in 4 Nvidia 2080Ti (11GB VRAM) GPUs and the batchsize is set to 8.
\subsubsection{Evaluation Metrics}
We evaluate our generated result in the iPER dataset from two aspects: self-imitation results with ground truth comparison and cross-imitation among different persons without ground truth. For self imitation, we mainly evaluate the quality of the reconstructed image with Structure Similarity\cite{wang2004image} (SSIM) and Learn Perceptual Image Patch Similarity\cite{zhang2018unreasonable} (LPIPS). SSIM measures image structure similarity, and LPIPS calculates the image patch's perceptual distances between generation result and ground truth. For cross imitation, Inception Score\cite{salimans2016improved} (IS) and Fréchet Inception Distance\cite{heusel2017gans} (FID) estimate the realness of generated images with machine perception. Meanwhile, all metrics are calculated in the DeepFashion dataset to evaluate its reality and quality.

\begin{table}
\caption{Quantitative results of motion transfer with various models on the iPER dataset. The asterisk (*) means that there is no need to evaluate the influence of pose normalization in self-imitation setting since the pose pair belong to the same person. The arrow near the metric name indicates the better results direction.}
\centering
\begin{tabular}{c|c|c|c|c}
\hline \multirow{2}{*} {} & \multicolumn{2}{|c|} { Self-Imitation } & \multicolumn{2}{c} { Cross-Imitation } \\
\cline { 2 - 5 } & SSIM $\uparrow$  & LPIPS $\downarrow$ & IS$\uparrow$ & FID $\downarrow$ \\
\hline 
FSV2V\cite{wang2019fewshotvid2vid} & 0.776 & 0.314  & \underline{2.36}  & 131.9 \\
LiquidGAN\cite{liu2019liquid} & 0.888 & 0.068  & 2.11  & 86.8 \\
GFLA\cite{ren2020deep}  & 0.909  & 0.059 & 2.15 & 128.6 \\
PoNA\cite{li2020pona}  & $\mathbf{0.972}$  & 0.075 & 2.27 & 139.2\\
PISE\cite{zhang2021pise}  & 0.910  & \underline{0.044} & 2.08 & \underline{80.3} \\
\hline
Baseline & 0.920  & 0.043 & 1.92 & 80.3\\
$w/i$ bc loss & 0.928  & 0.040 & 2.49 & 75.1\\
$w/i$ dual attn & 0.924  & 0.040 & 2.41 & 78.3 \\
Full & \underline{0.930}  & $\mathbf{0.036}$ & $\mathbf{2.58}$ & $\mathbf{72.9}$ \\
$w/o$ posenorm & *  & * & 2.19 & 84.1 \\
\hline
\end{tabular}
\label{tab:comparison_iper}
\end{table}

\begin{table}
\caption{Quantitative results of motion transfer with various models on the DeepFashion dataset. The arrow near the metric name indicates the better results direction.}
\begin{center}
\begin{tabular}{c|c|c|c|c}
\hline 
\cline { 2 - 5 } & SSIM $\uparrow$  & LPIPS $\downarrow$ & IS$\uparrow$ & FID $\downarrow$ \\
\hline 
LiquidNET\cite{liu2019liquid} & 0.696 & 0.470  & 3.47  & 28.1 \\
XingGAN\cite{tang2020xinggan} & 0.710 & 0.297  & 3.49  & 48.8 \\
GFLA\cite{ren2020deep}  & 0.701  & 0.221 & \underline{3.69} & 14.5\\
PoNA\cite{li2020pona}  & $\mathbf{0.775}$  & 0.406 & 3.33 & 32.3 \\
PINet\cite{zhang2020human}  & 0.648  & 0.216 & 3.41 & 15.2 \\
PISE\cite{zhang2021pise}  & 0.630  & \underline{0.2059} & 3.41 & \underline{13.6} \\
\hline
Baseline & 0.683  & 0.290 & 2.89 & 32.0 \\
$w/i$ bc loss & 0.709  & 0.220 & 3.65 & 12.8\\
$w/i$ dual attention & 0.715  & 0.237 & 3.52 & 16.9\\
Full & \underline{0.729}  & $\mathbf{0.187}$ & $\mathbf{3.77}$ & $\mathbf{8.4}$ \\
\hline
\end{tabular}
\label{tab:comparison_fashion}
\end{center}
\end{table}

\begin{figure*}[tb!]
\centering
\includegraphics[width=\linewidth]{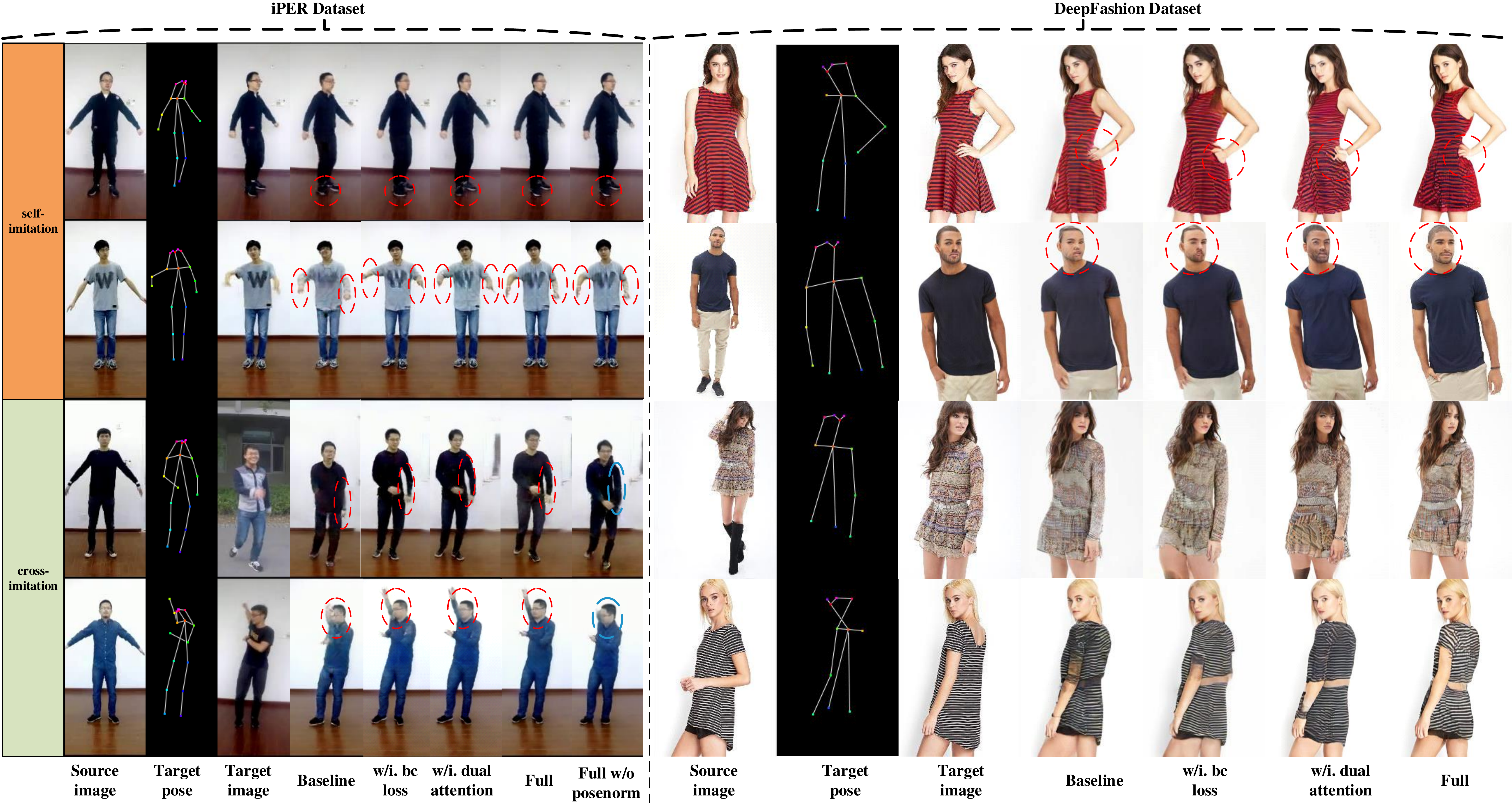}
    \caption{Qualitative results of ablation study in iPER and DeepFashion datasets.}
\label{fig:ablation_all}
\end{figure*}

\subsection{Comparison with the Conventional Methods}
\label{section:comparison}
\subsubsection{Qualitative Results}
\label{section:qualitative}
We visualize the generated results on iPER and DeepFashion datasets, compared with several state-of-the-art methods including FSV2V\cite{wang2019fewshotvid2vid}, LiquidNET\cite{liu2019liquid}, XingGAN\cite{tang2020xinggan}, PoNA\cite{li2020pona}, PINet\cite{zhang2020human}, and PISE\cite{zhang2021pise}. The FSV2V can only be employed on the video dataset, so we conduct its experiment on the iPER dataset. Besides, since XingGAN and PINet do not provide a pretrained model on the iPER dataset, we solely compare them on the DeepFashion dataset.

As shown in \autoref{fig:comparison_fashion}  and \autoref{fig:comparison_iper}, visualization results generated by state-of-the-art models are used for qualitative comparison to show the superiority of our method. Specifically, our model can generate high-fidelity images and maintain the human identity compared with FSV2V, XingGAN, and PoNA, while these methods fail to preserve the texture sharpness and image details. Moreover, even conditioned with complex source dressing patterns, our method can produce the most plausible garment details consistent with the source, eg., the sweater detail is preserved in our result (as shown in the fifth row of \autoref{fig:comparison_fashion}). Besides, our method can predict better texture and structure information in the occluded region than methods like GFLA, PINet, and PISE. For example, we achieve better size consistency of parts like bracelet and camisole (as circled in red in the first, second, and third rows of \autoref{fig:comparison_fashion}), as well as better details of shoes near the boundary (as shown in the second row of \autoref{fig:comparison_iper}). Additionally, the body shape and foot position in our results can remain the same as the source human. Although LiquidNET\cite{liu2019liquid} can roughly keep the output body shape invariant with the source, it is vulnerable to the inaccurate 3d model estimation result caused by huge camera coordinate difference. It shows a conflicting leg or head pose with the target (as shown in the third, fourth, and fifth rows of \autoref{fig:comparison_fashion}). More analysis will be presented in the Sec.\ref{section:effect_pn}. In summary, our method can guarantee the detailed source appearance recovery and maintain the body shape in the target pose simultaneously.

\subsubsection{Quantitative Results}
\label{section:quantitative}
We compare our method with several latest models on the iPER and DeepFashion datasets in \autoref{tab:comparison_iper} and \autoref{tab:comparison_fashion}, respectively. Specifically, to better evaluate the performance of our model in the video-based iPER dataset, we conduct cross-imitation evaluation where the source and target belong to different identities. We calculate the SSIM and deep feature-based LPIPS metric to assess the spatial similarity with ground truth using the generated images in the iPER dataset. Our SSIM is slightly lower than the PoNA\cite{li2020pona} because SSIM prefers more blurry images as pointed out by ~\cite{ledig2017photo, wang2018high, wang2018back} which indicates the inconformity  between higher SSIM score and better perceptual quality. We achieve the lowest LPIPS score, which is more consistent with human judgment. It means that our FDA-GAN can generate the images with better maintenance of perceptual structure similarity. In the cross-imitation setting, when there is no existing ground truth, we use IS and FID to measure the generation quality compared with worldwide images and source human image collections, respectively. Both metrics outperform the state-of-the-art methods, which validates our method's improvements in generation quality and realism. For the DeepFashion dataset, we can see that our model surpasses others in most metrics, which validates our method's improvement on the high-quality image generation.

\begin{figure*}[!tb]
	\centering
	\includegraphics[width=\linewidth]{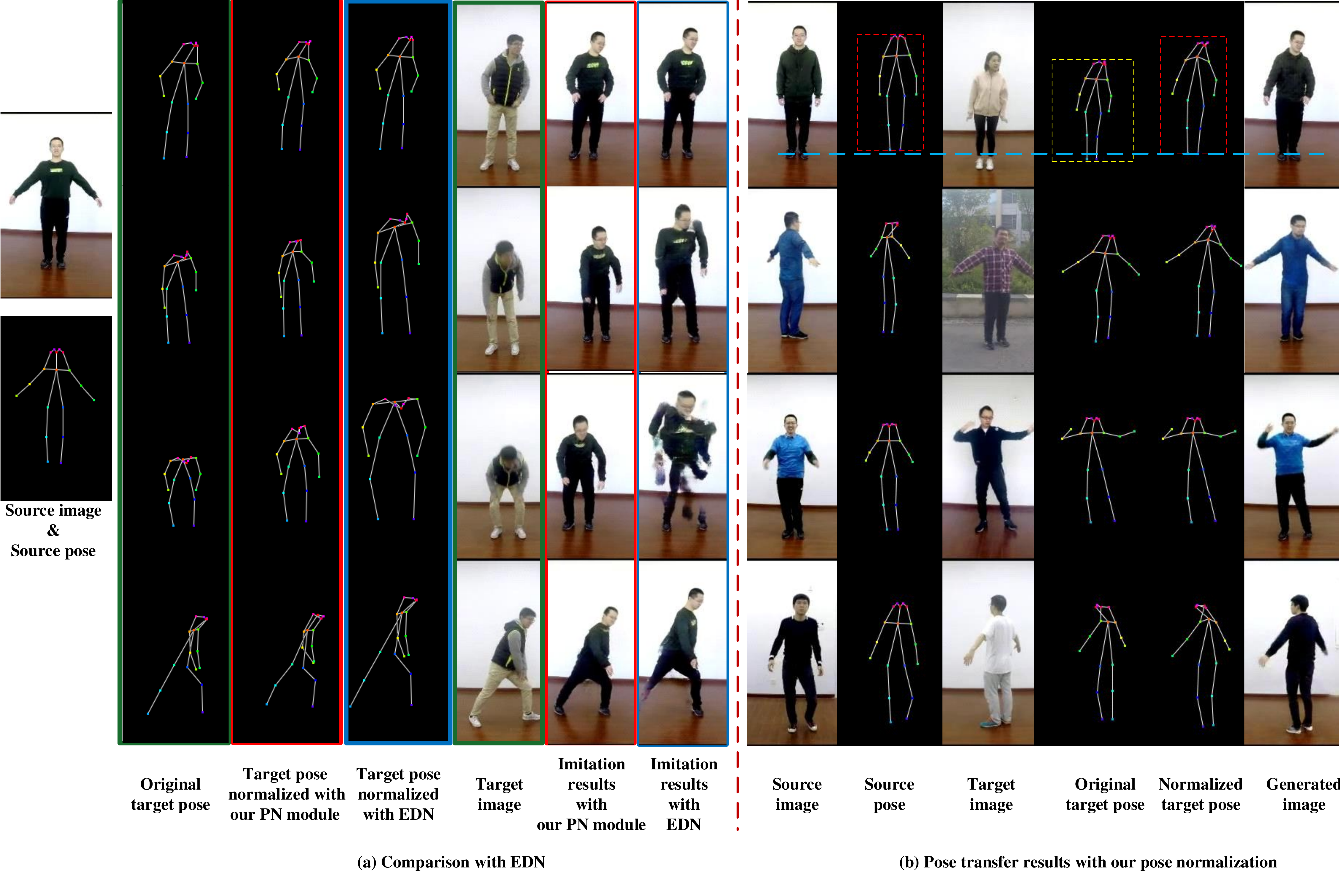}
	\caption{The analysis about our pose normalization module. (a) Comparison with previous pose normalization method in EDN\cite{chan2019everybody}. We generate images in different target poses with our PN Net and the EDN to qualitatively validate the effect of our normalization method. (b) Examples of pose transfer with our PN Net. We emphasize our PN Net's capacity to preserve source foot position with a horizontal blue line.}
	\label{fig:pnorm_show_and_comparison}
\end{figure*}

\begin{figure}[!htb]
	\centering
	\includegraphics[width=\linewidth]{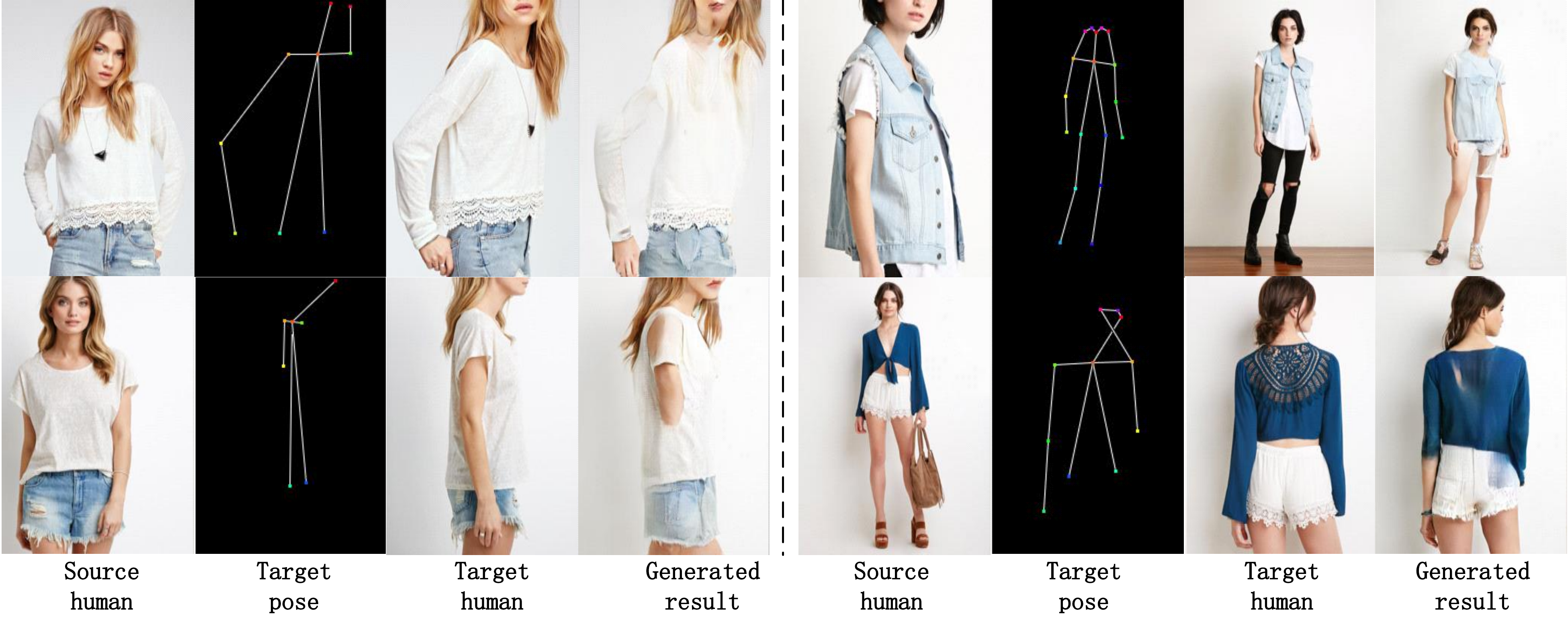}
	\caption{Examples of failure cases using FDA-GAN model.}
	\label{fig:failure}
\end{figure}

\begin{table}
	\caption{Quantitative results with varying search size $k$ and ${\Tilde{k}}$ of $top{\Tilde{k}}^2$ selection in our flow similarity attention module on the DeepFashion dataset. We highlight \underline{$\textbf{best}$}, $\textbf{second best}$ and \underline{third best} scores. ‘*’ and TC denote the default setting and the time cost for one forward process of our flow similarity attention module in our original manuscript, respectively.}
	\begin{center}
		\begin{tabular}{c|c|c|c|c|c}
			\hline 
			\cline { 1 - 6 }  Models & SSIM $\uparrow$  & LPIPS $\downarrow$ & IS$\uparrow$ & FID $\downarrow$ & TC(s) $\downarrow$\\
			\hline 
			k=10, ${\Tilde{k}}$=9 & \underline{0.725} & 0.205  & 3.59  & 9.7 & 0.460 \\
			k=10, ${\Tilde{k}}$=16 & 0.714 & 0.205  & 3.50  & 10.2 & 0.470 \\
			k=4, ${\Tilde{k}}$=4 & 0.714 & 0.196  & 3.60  & \underline{8.6} & \underline{$\mathbf{0.062}$} \\
			k=8, ${\Tilde{k}}$=4 & 0.723 & \underline{0.190}  & \underline{3.67}  & $\mathbf{8.5}$ & $\mathbf{0.296}$ \\
			k=16, ${\Tilde{k}}$=4 & \underline{$\mathbf{0.730}$} & \underline{$\mathbf{0.184}$}  & \underline{$\mathbf{3.79}$}  & \underline{$\mathbf{8.4}$} & 1.321 \\
			\hline\hline
			*k=10, ${\Tilde{k}}$=4 & $\mathbf{0.729}$ & $\mathbf{0.187}$  & $\mathbf{3.77}$  & \underline{$\mathbf{8.4}$} & \underline{0.456} \\

			\hline
		\end{tabular}
		\label{tab:hyperparameter}
	\end{center}
\end{table}

\subsection{Effectiveness of the FDA-GAN Framework}
\label{section:effect_model}
To demonstrate the effectiveness of our dual attention module and bidirectional consistency (bc) loss, we perform an ablation study with several variant models for comparison. 
\subsubsection{Baseline} The source feature maps are directly warped according to the feature flow map predicted by the flow estimator without bidirectional loss.
\subsubsection{w/i bc loss} This model adopts the same architecture as the baseline except for the bc loss constraint. Especially, to eliminate the influence from pose difference, we normalize the target pose for cross-imitation in the iPER dataset. 
\subsubsection{w/i dual attention} We propose this model to validate the efficacy of our dual attention module introduced in Sec.\ref{section:dualattn}. We still normalize the original target pose using our pose normalization network.
\subsubsection{Full model} We use the complete FDA-GAN, which contains all the modules.
\subsubsection{Full model w/o posenorm} We remove the pose normalization before the main FDA-GAN generator, which has dual attention and bc loss.

As shown in \autoref{tab:comparison_iper} and \autoref{tab:comparison_fashion}, the combination of bc loss and dual attention module prompts the result of all metrics. Compared with the baseline, the performance gain from bc loss proves that our forward-backward checking can help the flow estimator to calculate more accurate flow values. We observed that such benefit is positively correlated to the degrees of occlusion. In DeepFashion dataset, the degree of occlusion is higher due to the large amount of viewpoint changes. In iPER dataset, the viewpoint is almost the same for each person, leading to a relatively lower degree of occlusion. Regardless of the difference, the occlusion problem still exists, and that is the reason why BC loss has such effect on these two datasets. Furthermore, the dual attention module can sample more reasonable positions for target outputs. However, the direct warping baseline cannot supply correct source feature values, leading to degraded performance. We further tested the results of variant models trained with different search size $k$ and ${\Tilde{k}}$ of $top{\Tilde{k}}^2$ selection in our flow similarity attention module as shown in \autoref{tab:hyperparameter}, which indicates that our setting ($k$=10, ${\Tilde{k}}$=4) can achieve relatively good performance while maintaining appropriate training time consumption. The best setting ($k$=16, ${\Tilde{k}}$=4) requires about 2.9 times longer than our setting but the performance is not significantly improved. Our full FDA-GAN outperforms the other variants and achieves the best results.

Besides, from the visualization results in  \autoref{fig:ablation_all}, the results of these variants suffer blurry and unrealistic warped appearance generation in the face, shoe, or hand, as circled in red in \autoref{fig:ablation_all}. In contrast, the full model can generate more plausible results, intuitively discovered by the face and clothes details. It is worth noting that the clothes texture generated by our model is sharper and more similar to the source, which proves the effectiveness of our dual attention mechanism and bc loss. Particularly, the pose normalization merely works in the cross setting since it just improves the generation detail when the source differs from the target, as circled in blue in \autoref{fig:ablation_all}. More detailed analysis about pose normalization network is given in Sec.\ref{section:effect_pn}.

\subsection{Effectiveness of the Pose Normalization Network}
\label{section:effect_pn}
We propose the Pose Normalization Network (PN Net) to preprocess the target pose before the main FDA-GAN, which can tackle high variance in source and target pose pair and avoid degraded performance in pose transfer. To prove the effectiveness of a self-supervised trained pose normalization network, we generate a normalized pose map and then conduct motion transfer based on the normalized pose on the iPER dataset. Besides, we compare our PN Net with the normalization method proposed in EDN\cite{chan2019everybody}, which uses 2d scaling and translation to align the target pose.

To demonstrate superior performance in normalizing the poses, we generate person images based on our PN Net and the pose normalization method in EDN. As shown in \autoref{fig:pnorm_show_and_comparison}(a), when the human is in an upright posture, the generated results by both methods show no noticeable difference (the first row of the \autoref{fig:pnorm_show_and_comparison}(a)), which indicates that our method has a similar effect when the bone length changes in the approximate 2d plane. However, in cases like bending or leg pressing, joint positions move in the 3d space. EDN fails to preserve the source pose and adapt the original target joint to the right positions when there is high variance between the source and target, as shown in the \autoref{fig:pnorm_show_and_comparison}(a). The simply scaling and translation way cannot deal with the complex body structure deformation and would cause misalignment w.r.t joint positions. We might have to introduce a complicated linkage mechanism with many handcrafted hyperparameters if we implement 2d pose normalization manually. In contrast, our method uses a self-supervised model to learn how to transfer from one body structure to another and achieve better performance.

Notably, from the visualization results of \autoref{fig:pnorm_show_and_comparison}(b), we can intuitively discover that the normalized pose can keep the action pattern unchanged (the action he/she is doing) and adapt the original pose to align with the source human's body size and global position.

\section{Discussion}
\label{section:discussion}
\subsection{Limitations}
Despite the improvement over previous methods of our model, we have to point out that the generation of occlusion regions remains a challenge in pose transfer.  \autoref{fig:failure} presents some failure cases with our FDA-GAN model. Our model generates the occluded regions by searching regions with similar flow values, which is less useful when facing large occlusion and may result in unreasonable and blurry texture in these areas. Besides, the errors of pose estimation can be another big issue since the model cannot acquire reasonable pose guidance. 

\subsection{Future improvements}
In the case with large occlusion, which means that the target generation cannot get much help from the source, the pose transfer task can be more like the unconditional image generation and need more common sense reasoning. So in the future, we seek to enhance the capability of the unconditional generation to solve this problem. To alleviate the negative effect of pose estimator, we may try to get the latent pose guidance from the target image directly without reliance upon off-the-shelf pose estimation.

\section{Conclusion}
\label{section:conclusion}
In this paper, we have proposed a novel method to perform human pose transfer. It augments the flow generation module with bidirectional consistency and employs a dual attention module to integrate the source and target features better. Furthermore, to remove the impact of disparate skeleton structure and global position between source and target, a pose normalization network has been trained to generate consistent pose skeletons. Extensive experiments qualitatively and quantitatively prove that our model outperforms others.


%

\ifCLASSOPTIONcaptionsoff
  \newpage
\fi


\bibliographystyle{IEEEtran}
\bibliography{IEEEabrv,ref}

%

\end{document}